# A PATTERN-MINING DRIVEN STUDY ON DIFFERENCES OF NEWSPAPERS IN EXPRESSING TEMPORAL INFORMATION

Yingxue Fu[1, 2] and Elaine Uí Dhonnchadha[2]

[1]School of Computer Science, University of St Andrews, Scotland, UK
[2] Center for Language and Communication Studies, Trinity College
Dublin, Dublin 2, Ireland

## ABSTRACT

*This paper studies the differences between different types of newspapers in expressing temporal information, which is a topic that has not received much attention. Techniques from the fields of temporal processing and pattern mining are employed to investigate this topic. First, a corpus annotated with temporal information is created by the author. Then, sequences of temporal information tags mixed with part-of-speech tags are extracted from the corpus. The TKS algorithm is used to mine skip-gram patterns from the sequences. With these patterns, the signatures of the four newspapers are obtained. In order to make the signatures uniquely characterize the newspapers, we revise the signatures by removing reference patterns. Through examining the number of patterns in the signatures and revised signatures, the proportion of patterns containing temporal information tags and the specific patterns containing temporal information tags, it is found that newspapers differ in ways of expressing temporal information.*

## 1. INTRODUCTION

Newspapers are broadly categorized into two types: broadsheets and tabloids. As different newspapers target different audiences, the newspapers may use different words and sentence structures in their reports and thus have distinctive styles. Meanwhile, since newspapers aim at reporting events in a timely manner and things evolve with time, newspapers typically contain more temporal information than the other kinds of texts. Although the stylistic differences of newspapers have been studied thoroughly [2, 3, 4, 5], how newspapers differ in expressing temporal information remains an under-explored topic. This may be attributed to the simplified view of temporal information in natural language texts. The research on automatic processing of temporal information, however, shows the complex nature of temporal information.

Therefore, it would be of some interest and significance to investigate whether different newspapers express temporal information in different ways. As temporal information can be annotated automatically with temporal processing tools developed by the research community, it is possible to incorporate temporal information into the syntactic analysis of news articles.

Tabloid articles tend to be shorter and less serious in content compared to broadsheets. Therefore, it may be legitimate to presume that tabloids contain less temporal information than broadsheets and that they avoid using explicit temporal expressions. However, evidence supporting this assumption has not been found.

To investigate this question, we create a corpus of newspaper articles annotated with temporal information so that sequences containing temporal information tags can be extracted. Inspired by previous research [1], we use the method of mining part-of-speech (POS) skip-gram patterns from the sequences and deriving signatures for the different newspapers.

Skip-gram modeling [6] is a technique proposed to solve the problem of data sparsity in natural language processing (NLP). Skip-grams are sequences of tokens that are similar to fixed-length n-grams but allow a gap of a user-defined size between adjacent tokens, for example, the pattern "hit ball" can be extracted from the sentence "I hit the tennis ball" with a gap of two.

Although the parameter setting of the pattern mining algorithm and the steps of deriving signatures of the newspapers are similar to previous research [1], the purpose of our research is different. This method [1] has been used to test the effectiveness of the POS skip-gram patterns in the task of authorship attribution, in other words, to test if the extracted POS skip-gram patterns can be used as a stylistic feature to characterize an author's work. In contrast, our focus is to compare the skip-gram patterns containing temporal information tags in the signatures of different newspapers so that a better understanding about how newspapers differ in expressing temporal information can be obtained. As research shows that the POS skip-gram patterns that form the signature of an author are effective in capturing the style of an author, we delve into the signatures of the newspapers and pay attention to the patterns that are formed by temporal information tags to investigate our research question.

Contrary to our preconception, our analysis shows that the Daily Mirror, generally described as a tabloid, contains a greater proportion of temporal information in its signature and the temporal information tends to be expressed with explicit temporal expressions, which differs from the Guardian, which is typically described as a broadsheet.

It is worth mentioning that in spite of the wide application of neural networks in natural language processing tasks, considering our research question, the neural approach is a less desirable option because of the difficulty in observing the process and interpreting the results.

The next section introduces related work in the fields of temporal processing, pattern mining and the application of pattern mining for authorship attribution. Section 3 discusses the methodology, including the details of corpus creation and annotation, the algorithm for pattern mining and the steps for obtaining the signatures of the newspapers. Section 4 presents the results and discussion. Section 5 concludes the paper and points out future work.

## 2. RELATED WORK

To create the corpus annotated with temporal information, it is necessary to understand the techniques for processing temporal information in NLP. A review of the research on automatic extraction of temporal expressions and temporal relations and automatic identification of events is presented.

### 2.1. Automatic Extraction of Temporal Information

Generally speaking, temporal information in natural language texts can be embodied in three ways: a) temporal expressions: at 10:30, on Christmas Day, recently and the like; b) tense and aspect of verbs, such as goes, went, had gone, is eating, has been eating; c) temporal relations, for instance, the explosion happened soon after he got out of the theater.

The annotation of temporal information has been standardized under the ISO-TimeML scheme [7]. EVENT, TIMEX3, SIGNAL and LINK are the four major tags in this annotation scheme.
EVENT denotes things that happen or occur and may be related to temporal expressions or involve temporal relations, for example, "the car crash" in the sentence "he was killed in the car crash yesterday" is an event associated with the temporal expression "yesterday".

TIMEX3 marks up explicit temporal expressions which may have the attributes of "duration", "date", "time" and "set". For example, in the sentence "the rain lasted for two weeks", "two weeks" has the attribution of "duration"; in the sentence "George was born on December 12, 1979", the temporal expression "December 12, 1979" has the attribute of "date"; "three years ago" in the sentence "he left the village three years ago" has the attribute of "time"; and in the sentence "I visited her twice a week that year", "twice" has the attribute of "set".

SIGNAL is used to annotate function words which reveal the connection between temporal objects, such as "before", "during", and "when".

LINK is a general tag for temporal relations. The annotation of temporal relations depends on the extraction and annotation of temporal expressions and events. Under the TimeML annotation scheme, LINK can be divided into three types:

- TLINK represents temporal relations between events or between an event and a temporal expression, which are rooted theoretically in the 13 temporal relationships [8], e.g. "before", "equal", "meet", "overlap", "during", "start" and "finish";
- SLINK represents a subordinate relationship between two events or between an event and a signal, where a introducing relation is typically present, for example, the relation between "wanted" and "leave" in the sentence "Mary wanted John to leave his family", and the relation between "regret" and "wear" in "Mary regrets that she didn't wear high heels that day";
- ALINK represents the relationship between an aspectual event and its argument event, such as "stop talking", "keep reading" and "starts to rain" [9].

Generally speaking, temporal information extraction is implemented using rule-based methods, machine learning techniques, or a hybrid of the two [10]. HeidelTime [11] and SUTime [12] are two of the best performing tools for temporal expression extraction, and both use rule-based methods which make normalization of temporal expressions easier. HeidelTime is the best performing system in TempEval-2 (http://semeval2.fbk.eu/semeval2.php?location=tasksT5) for extracting temporal expressions. Each temporal expression is viewed as a three-tuple consisting of a temporal expression, the type of the temporal expression (i.e., one of four types: date, time, duration and set), and the normalized value of the temporal expression. The goal is to extract the temporal expression and assign the type and calculate the normalized value correctly. SUTime is a rule-based temporal tagger built on regular expression patterns. Three types of rules are applied: text regex rules which are applied first to map simple regular expressions over characters or tokens to temporal representations; compositional rules that map regular expressions over chunks that are formed by tokens and temporal objects to temporal representations, (these rules being applied iteratively after the text regex rules); and filtering rules which discard ambiguous expressions that are likely to be non-temporal expressions, e.g. rules designed for polysemous words, such as "fall" [12]. The dependence on patterns for extraction of temporal expressions suggests a close connection between patterns and temporal expressions.

Machine learning techniques have also been proposed as a method of inferring the temporal relation linking a main clause and a subordinate clause attached to it. This is an example of learning temporal relations with machine learning techniques. There are also models [14, 15] which use hybrid methods. As indicated by [10], both rule-based approaches and systems implemented using machine learning algorithms typically rely on grammatical and syntactical attributes, such as POS tags, tense and so on.

Due to the dependence of techniques for automatic processing of temporal information on syntactic patterns of texts, we mix POS tags with temporal information tags for analyzing the stylistic features of newspaper articles.

## 2.2. Research on Pattern Mining

The second major part of our theoretical background is pattern mining. Pattern mining is one of the fundamental tasks of data mining and it consists of finding interesting, useful or unexpected patterns in a database [16]. This field originates from research [17] that provides an algorithm for solving the problem of discovering patterns of items bought by customers at a store, which may be used for commercial purposes, such as product catalogue design, add-on sales, store layout and customer segmentation based on purchasing patterns.

Under the framework of data mining, the temporal information in a dataset is used in the data preparation step for ordering items for subsequent pattern mining. In this case, the temporal aspect is not assigned with much weight. In the case of mining meaningful patterns from the database of customer purchase records, each transaction is considered as an itemset and the transactions are sorted based on the time of the transactions before a pattern mining algorithm is applied [18].

Temporality in pattern mining becomes an important consideration under the framework of temporal pattern mining. It is noticed that the correlation degree between some terms or patterns can change with time. [19] illustrates this point with the pattern formed by "Hillary Clinton" and "Candidate" which are correlated more strongly in 2008 than the pattern formed by "Hillary Clinton" and "Secretary of State", while "Hillary Clinton" and "Secretary of State" form a more prominent pattern in 2009 than the pattern formed by "Hillary Clinton" and "Candidate". The length of time interval is considered as an important factor in the prediction task [20] and the pattern discovery task [21]. [22] point out a problem with vanilla sequential pattern mining that does not take time gaps into account: if most of the customers buy B after A, and C after B, the manager can use this pattern to promote B when a customer purchases A and promote C when a customer buys B. However, if the time intervals between the purchases are not known, improper product recommendation would occur. This happens when customers purchase B several days after A, and buy C a certain amount of time after buying B, rather than buying B immediately after A and buying C with the same time gap after purchasing A. To solve this problem, a time-gap sequential pattern mining algorithm is proposed in their paper.

The above study demonstrates how the temporal aspect is handled in data mining. As temporal expressions are generally not expressed by a single word, the phrase is a more suitable unit for study. Events and temporal relations can be treated as patterns in a text. When temporal patterns are added in pattern mining, some interesting patterns may be obtained. However, the combination of temporal information and syntactic patterns in texts has not yet explored.

## 2.3. Pattern Mining for Authorship Attribution

Just as pattern mining algorithms can be used to identify the customers' purchasing habits, they can also be used for characterizing the distinct writing styles of different authors. [23] apply pattern mining techniques to e-mail forensic analysis. As suggested by the authors, the two most widely used machine learning algorithms for authorship attribution, Decision Trees and Support Vector Machines (SVM), have limitations in forensic investigations. When a decision tree is built, a decision node is constructed based only on the local information of one attribute, and the combined effect of several features is not captured. A second drawback of using Decision Trees in this task is that the same set of attributes is used for all the suspects, which could be manipulated for supporting wrong arguments. The problem with the SVM is that it is a learning function that works like a black box whose result is difficult to interpret for forensic purposes. This property makes SVM a less desirable choice.

Therefore, the authors present a method based on a pattern mining algorithm to extract the unique write-print of each suspect. Write-print is a term which denotes the patterns that can uniquely capture the writing style of an individual. The Apriori algorithm [17] is adopted in the pattern mining step. Only a pattern that appears above a pre-set threshold frequency, which is defined as 'support', is considered to be frequent. The support value is calculated as the percentage of emails that contain the pattern in the training set of a suspect author. The second step is to filter out common frequent patterns so that a pattern in the write-print of one suspect does not appear in the write-print of another. A write-print perfectly matches a test email if the test email contains every pattern in the write-print. Since frequent patterns may vary in occurrence frequency, and the more frequent patterns are generally more important than the others, the support of a frequent pattern is taken into account in the function for calculating the similarity between the write-print and the test email. This paper demonstrates the advantage of using pattern mining algorithms for forensic use over using machine learning algorithms.

The use of frequent POS skip-grams for authorship attribution is explored in [1]. The basic idea of this paper is to combine POS skip-grams and a top-k sequential pattern mining algorithm for the authorship attribution task. POS skip-grams are constructed in a similar way as POS n-grams except for the fact that a gap is allowed between adjacent POS tags, thus introducing an additional parameter representing the size of the gap. With the top-k sequential pattern mining algorithm, only the most frequent POS skip-grams are considered. The data for the experiments, comprising 30 books written by 10 authors, is taken from the Gutenberg Project (https://www.gutenberg.org). Each author is represented by three texts. Since it is believed that an author not only writes in his own style but also shares common patterns with the other authors, in order to obtain the unique signature of the author, patterns in the union of the other authors , i.e. reference patterns, will be removed from the initially obtained signature of the author. The Pearson correlation coefficient is chosen to measure the correlation between an anonymous test text and the signature of each author. It is shown that using POS skip-grams provides better performance than using POS bigrams and trigrams. The influence of the parameters of the top-k pattern mining algorithm on the overall performance and on the classification accuracy for each author is also studied.

It can be seen from the above papers that applying pattern mining algorithms for authorship attribution typically includes the steps of extracting more representative patterns from a text, finding the unique patterns of each author, and using a function for calculating the correlation/similarity between a test file and the set of unique patterns of each author which is called a write-print or a signature. Compared with the models for authorship attribution implemented using machine learning algorithms, the approach based on pattern mining has some advantages, such as being capable of discovering unique patterns for each author, which can serve as more credible evidence in forensic scenarios. As our research question requires the analysis of the patterns containing temporal information tags, the research on applying pattern mining techniques for authorship attribution can be a source of inspiration.

## 3. METHODOLOGY

To study the stylistic differences between different newspapers, a corpus is created by the authors.

### 3.1. Corpus Creation

The first step is to create a corpus annotated with temporal information. 1200 news articles are collected from four online newspapers: BBC (https://www.bbc.com), the Guardian (https://www.theguardian.com/uk), the Independent (https://www.independent.co.uk), and the Daily Mirror (https://www.mirror.co.uk). To reduce the influence of text categories on the result, texts classified under the categories of sports, politics and science & technology are collected in the same number. The publishing time of the news articles ranges from January 2020 to May 2020.

### 3.2. Corpus Annotation

The TARSQI Toolkit (TTK) is a suite of temporal processing modules for automatic temporal and event annotation of natural language texts [24]. TTK allows multiple linguistic annotation tasks to be performed, including tokenization, lemmatization, chunking, POS tagging, sentence and phrase boundary detection, temporal expression annotation and temporal relation annotation. It integrates several modules for temporal processing, including: the PreProcessor for tokenization, POS tagging and chunking, which is actually implemented by the TreeTagger [25]; GUTime for extracting temporal expressions; Evita for extracting events; Slinket for modal parsing; S2T for temporal repercussions of modal relations; Blinker for opportunistic pattern-based parsing of temporal relations; Classifier which is a MaxEnt classifier trained on the TimeBank corpus for identifying temporal relations between previously recognized events and temporal expressions in a text [24]; and Link Merger for ensuring consistency of all the temporal relations.

As the latest release of TTK is mainly written in Python 2 which has been declared End of Life in 2020, the source code of TTK released on GitHub cannot run without manual correction and not all the modules can work normally. Based on the tagger's performance in the pilot experiment, only modules that work are selected. Hence, PreProcessor, GUTime, Evita, Slinket and S2T are used in the annotation process. The statistics of the corpus is presented below.

Table 1. Statistics of the corpus.

|                | BBC    | The Guardian | The Independent | The Daily Mirror |
|----------------|--------|--------------|-----------------|------------------|
| <s>            | 10262  | 11013        | 8627            | 8438             |
| <lex>          | 218498 | 254613       | 196453          | 175833           |
| <vg>           | 18316  | 21696        | 16679           | 14686            |
| <ng>           | 20127  | 23175        | 18082           | 15843            |
| <EVENT>        | 14946  | 17612        | 13750           | 12073            |
| <TIMEX3>       | 2487   | 2501         | 1957            | 2004             |
| <SLINK>        | 753    | 764          | 537             | 560              |
| <TLINK>        | 674    | 667          | 454             | 492              |
| <ALINK>        | 0      | 0            | 0               | 0                |

It can be seen that ALINK is not recognized, which may be attributed to the exclusion of some modules of TTK. The meaning of most of the tags in Table 1 has been explained in section 2.1. Apart from the typical tags specified under ISO-TimeML annotation scheme, such as <EVENT>, <TIMEX3>, <SLINK>, <TLNK> and <ALINK>, there are other tags generated by the PreProcessor of TTK: <s> represents the marker of a sentence (without the closing tag </s>); <lex> denotes tokens; <vg> means verb phrase; and <ng> represents noun phrase.

From the statistics, it can be seen that the Guardian ranks the first in terms of the number of sentences, followed by BBC, the Independent and the Daily Mirror. Articles from the Guardian, which is classified as a broadsheet, are longer than the others, and articles from the Daily Mirror, which is generally described as a tabloid, are the shortest of all. The same trend can be found in the statistics of tokens, verb phrases, noun phrases and EVENT, which may be explained by the fact that these aspects are closely related to the low-level linguistic tasks, such as POS tagging. As to the numbers of temporal expressions represented by <TIMEX3> and temporal relations including <SLINK> and <TLNK>, the Daily Mirror exceeds the Independent.

### 3.3. The TKS Algorithm

The Top-K Sequential (TKS) pattern mining algorithm is used in the pattern mining step because it outperforms TSP which is the current state-of-art algorithm for the same task by more than an order of magnitude in terms of execution time and memory usage [26].

In the field of sequential pattern mining, the task of finding the most frequent sequential patterns is associated with the question of how to define the threshold value for being "frequent" so as to obtain an appropriate number of patterns. If too many patterns are discovered, the patterns might be less representative of the data and the computational costs are high both for the algorithm and further processing, while if too few patterns are found, some interesting or important patterns might be missed.

Therefore, to reduce the difficulty of the problem, the question of mining the most frequent sequential patterns is redefined as mining the top-k sequential patterns, where k is a user-defined number of sequential patterns to be discovered.

Before the details of the TKS algorithm are explained, some concepts may have to be clarified. A dataset can be formally defined as $S=\{s_1, s_2, s_3, \ldots s_i\}$, where $s_1...s_i$ are sequences. A set of items I may be defined as $I=\{i_1, i_2, i_3,...i_m\}$ and an itemset t is a set of items that belong to I, such as $\{i_1\}$ or $\{i_2, i_3\}$. Each sequence may contain one or more itemsets, for instance, $s_1 =\{t_1, t_2, t_3\}$. A k-item sequence means a sequence $s=\{t_1, t_2, t_3, \ldots t_k\}$, where $t_n$ is an itemset for $1 \leq n \leq k$. The *support* denotes the number of sequences in S that contain a specific pattern. It can also be expressed as the ratio of sequences that contain the pattern with respect to the total number of sequences in the database.

Each sequence can be considered either as a sequence-extended sequence or an itemset-extended sequence. Sequence-extension, which is also referred to as s-extension, means generating a new pattern by appending a new itemset after the existing itemsets of a sequence. For example, for $s_1=(\{a\}, \{b\}, \{c\})$ and $s_2=(\{a\}, \{b\}, \{c\}, \{d, e\})$, $s_2$ is an s-extension of $s_1$. Itemset-extended sequence, which is also called i-extension, means generating a new pattern by adding a new item to the last itemset of a sequence. For instance, for $s_1=(\{a\}, \{b\}, \{c\})$ and $s_2=(\{a\}, \{b\}, \{c, d, e\})$, $s_2$ is an i-extension of $s_1$.

The TKS algorithm employs a vertical database representation and the basic candidate-generation procedure of SPAM [27]. The meaning of vertical database representation may be understood in this way: given a database with *m* items and *s* sequences, each sequence may be identified with a unique ID and each of the *m* items may be represented separately by its presence in the itemsets of the sequence. Table 2 gives an illustration of horizontal database representation from which the vertical database representation may be derived.

Table 2.  An example of horizontal database representation.

| SID | Sequence |
|-----|----------|
| 1   | ({a, b}, {c}) |
| 2   | ({a, c}, {a, d}) |
| 3   | ({c, d}) |

In Table 2, SID denotes the id of each sequence in the database, and the unique items are {a, b, c, d}. In the sequence with the sequence ID 1, *a* appears in itemset 1, *b* appears in itemset 1, *c* appears in itemset 2, and *d* does not appear in any itemset. When the above table is turned into vertical representation, four tables will be generated so that each of the items in {a, b, c, d} is represented by its presence in the itemsets of the respective sequence:

Table 3.  Vertical representation of *a*.

| SID | Itemsets |
|-----|----------|
| 1   | 1 |
| 2   | 1, 2 |
| 3   |  |

As can be seen from Table 3, item *a* appears in the first itemset in sequence 1 and the first and second itemsets in sequence 2 but does not appear in sequence 3.

Table 4.  Vertical representation of *b*.

| SID | Itemsets |
|-----|----------|
| 1   | 1 |
| 2   |  |
| 3   |  |

As indicated by Table 4, item *b* only appears in the first itemset in sequence 1.

Table 5.  Vertical representation of *c*.

| SID | Itemsets |
|-----|----------|
| 1   | 2 |
| 2   | 1 |
| 3   | 1 |

As indicated by Table 5, item *c* appears in the second itemset in sequence 1, the first itemset in sequence 2 and the first itemset in sequence 3.

Table 6.  Vertical representation of *d*.

| SID | Itemsets |
|-----|----------|
| 1   |  |
| 2   | 2 |
| 3   | 1 |

As indicated by Table 6, item *d* appears in the second itemset in sequence 2 and the first itemset in sequence 3.

The database is scanned only once to obtain the vertical database representation and calculate the support of each item. Starting with the items, candidate patterns obtained through s-extension and i-extension are searched. If the support of a candidate pattern generated in this way surpasses a pre-set threshold, the candidate pattern will be used as basis for generating further candidate patterns through s-extension and i-extension. Infrequent patterns will not be extended to form frequent patterns, which is called the Apriori property [26].

TKS and SPAM are similar in terms of the vertical database representation and basic candidate generation procedure described above. However, TKS redefines the frequent pattern mining problem as discovering top-k sequential patterns and new strategies are introduced.

The initial threshold is set to 0. Then the basic candidate generation procedure is applied. When a pattern is found, it is added to the list of patterns which are ordered based on the supports of the patterns. When *k* sequential patterns are found, the threshold is raised to the support of the pattern with the lowest support in the list, so that patterns with supports lower than the threshold will not be considered. The process continues until no more patterns can be found. In this way, the problem of mining the most frequent sequential patterns is turned into the task of mining the top-k sequential patterns.

The second strategy is to extend the most promising sequential patterns first [26]. This strategy means that among the set of patterns that can be extended to form new patterns, the pattern with the highest support is extended first. In this way, the most promising patterns are found first and the threshold will be increased faster, thereby improving the efficiency of the algorithm.

The third strategy is to discard infrequent items in the generation of candidate patterns [26]. With the increase of the threshold, the items whose supports are below the threshold are not considered and if a sequence contains a single infrequent item, the item will be recorded in a hash table and skipped when patterns are extended.

A special structure called a precedence map (PMAP) is introduced. Its basic form is <j, n, s> for s-extension and <j, n, i> for i-extension. For example, if an item has PMAP <e, 3, s>, it means that the item is followed by e in three sequences of the database by means of s-extension.

The application of the TKS algorithm for pattern mining is implemented using the Sequential Pattern Mining Framework (SPMF), which is an open-source data mining library offering implementations of more than 55 data mining algorithms [28]. Compared with other open source data mining libraries such as Weka (https://www.cs.waikato.ac.nz/ml/weka/), Mahout (http://mahout.apache.org/) and Knime ( http://www.knime.org/), SPMF specializes in frequent pattern mining.

### 3.4. The Implementation

As the files tagged with TTK are saved as xml files, the xml.etree.ElementTree module (https://docs.python.org/3/library/xml.etree.elementtree.html) is used for parsing the files.

```
</source_tags>
<tarsqi_tags>
  <docelement begin="2" end="4219" id="d1" origin="DOCSTRUCTURE" type="paragraph" />
  <s begin="2" end="110" id="s1" origin="PREPROCESSOR" />
  <lex begin="2" end="4" id="l1" lemma="hm" origin="PREPROCESSOR" pos="ITJ" text="HM" />
  <lex begin="5" end="12" id="l2" lemma="revenue" origin="PREPROCESSOR" pos="NN1" text="Revenue" />
  <lex begin="13" end="16" id="l3" lemma="and" origin="PREPROCESSOR" pos="CJC" text="and" />
  <lex begin="17" end="24" id="l4" lemma="custom" origin="PREPROCESSOR" pos="NN2" text="Customs" />
  <lex begin="25" end="26" id="l5" lemma="(" origin="PREPROCESSOR" pos="PUL" text="(" />
  <lex begin="26" end="30" id="l6" lemma="<unknown>" origin="PREPROCESSOR" pos="NP0" text="HMRC" />
  <lex begin="30" end="31" id="l7" lemma=")" origin="PREPROCESSOR" pos="PUR" text=")" />
  <vg begin="32" end="42" id="c1" origin="PREPROCESSOR" />
  <lex begin="32" end="34" id="l8" lemma="be" origin="PREPROCESSOR" pos="VBZ" text="is" />
  <lex begin="35" end="42" id="l9" lemma="engage|engaged" origin="PREPROCESSOR" pos="VBN" text="engaged" />
  <EVENT begin="35" end="42" aspect="NONE" class="OCCURRENCE" eid="e1" eiid="ei1" epos="VERB" form="engaged" origi
  <lex begin="43" end="45" id="l10" lemma="in" origin="PREPROCESSOR" pos="PRP" text="in" />
  <ng begin="43" end="45" id="c2" origin="PREPROCESSOR" />
  <lex begin="46" end="47" id="l11" lemma="a" origin="PREPROCESSOR" pos="AT0" text="a" />
  <lex begin="48" end="54" id="l12" lemma="bitter" origin="PREPROCESSOR" pos="AJ0" text="bitter" />
  <lex begin="55" end="58" id="l13" lemma="war" origin="PREPROCESSOR" pos="NN1" text="war" />
  <lex begin="59" end="61" id="l14" lemma="of" origin="PREPROCESSOR" pos="PRF" text="of" />
  <lex begin="62" end="67" id="l15" lemma="word" origin="PREPROCESSOR" pos="NN2" text="words" />
  <lex begin="68" end="72" id="l16" lemma="with" origin="PREPROCESSOR" pos="PRP" text="with" />
  <ng begin="68" end="72" id="c3" origin="PREPROCESSOR" />
  <lex begin="73" end="76" id="l17" lemma="mp|mps" origin="PREPROCESSOR" pos="NN2" text="MPs" />
  <lex begin="77" end="81" id="l18" lemma="over" origin="PREPROCESSOR" pos="PRP" text="over" />
  <ng begin="77" end="81" id="c4" origin="PREPROCESSOR" />
  <lex begin="82" end="85" id="l19" lemma="the" origin="PREPROCESSOR" pos="AT0" text="the" />
  <lex begin="86" end="95" id="l20" lemma="so-called" origin="PREPROCESSOR" pos="AJ0" text="so-called" />
  <lex begin="96" end="97" id="l21" lemma="'" origin="PREPROCESSOR" pos="PUQ" text="'" />
  <lex begin="97" end="101" id="l22" lemma="loan" origin="PREPROCESSOR" pos="NN1" text="loan" />
  <lex begin="102" end="108" id="l23" lemma="charge" origin="PREPROCESSOR" pos="NN1" text="charge" />
  <lex begin="108" end="109" id="l24" lemma="'" origin="PREPROCESSOR" pos="PUQ" text="'" />
  <lex begin="109" end="110" id="l25" lemma="." origin="PREPROCESSOR" pos="." text="." />
```

Figure 1.  Illustration of an annotated file

Fig. 1 shows a part of an annotated file. To keep the original structure of the annotation and the syntactic patterns generated by the PreProcessor of TTK, the principle for extracting the tags is to keep the tags of <s>, <ng>, <vg>, <EVENT> and <TIMEX3>, extract the POS tag of the corresponding <lex>, and keep the stop as "." in the output sequence. The stop is required by SPMF so that the sentences can be treated as different sequences rather than treating the whole text as one sequence, which might cause misleading results. As <EVENT> is generally associated with a noun or verb which is important for maintaining the grammaticality of the text, the method of extracting <EVENT> is to extract both the tag of <EVENT> and the POS tag of the token annotated with <EVENT>. The sequences comprised by the above tags extracted from each news article are saved into separate files with the required extension of ".text" so that SPMF can recognize the file as a text document.

The following sentence is taken from a news article from BBC: "Team Ineos have withdrawn from all races until 23 March following the death of sporting director Nico Portal and the "very uncertain situation" surrounding the coronavirus outbreak." (Original text: https://www.bbc.com/sport/cycling/51737966).

The corresponding sequence extracted from the annotated file is as follows: "s NN1 NN2 VBB VBN vg EVENT VBN PRP ng DT0 NN2 PRP ng TIMEX3 ng CRD NP0 VBG vg EVENT VBG AT0 NN1 PRF AJ0 NN1 NP0 NP0 CJC AT0 PUQ AJ0 AJ0 NN1 PUQ VBG vg EVENT VBG AT0 NN1 NN1". (See a complete list of POS tags at http://ucrel.lancs.ac.uk/bnc2/bnc2guide.htm)

The second step is to extract signatures for BBC, the Guardian, the Independent, and the Daily Mirror. As each news article is independent, their top-k POS skip-gram patterns are extracted separately. The TKS algorithm has the following parameters: the number $k$ of POS skip-gram patterns to be found for each article, the minimum pattern length *minlen*, the maximum pattern length *maxlen*, and the maximum gap between the POS tags $g$ (when $g$ is set to zero, the skip-grams are the same as fixed-length n-grams). Based on the previous experiment results [1], the parameters are set as follows: $k$=250, *minlen*=1, *maxlen*=2 and $g$=1. The extracted patterns are saved separately for each news article.

To test the performance of the authorship attribution method, 75% of the files containing the extracted POS skip-gram patterns of each newspaper are used as training data and 25% are used as test data. The initial signature of a newspaper is formed by the patterns that appear in the extracted patterns of all or the majority of the news articles of the newspaper. For example, the initial signature of BBC is obtained by calculating the intersection of the top-250 POS skip-gram patterns of 225(=300*0.75) news articles.

The POS skip-gram patterns of a newspaper are obtained by concatenating the extracted top-250 POS skip-gram patterns of all the 225 news articles of the newspaper. Then pandas.dataframe API (https://pandas.pydata.org/pandas-docs/stable/reference/api/pandas.DataFrame.html) is used to read the files and find the duplicate patterns from the concatenated file, which are the patterns that the 225 files of patterns have in common. Because there are no repetitive patterns in a single file, if a pattern appears 225 times in the concatenated file, it must be a pattern that the 225 files of extracted patterns have in common.

Authors write in their own styles but they may share common structures with the other authors writing in the same language. To make sure that the signature obtained for each newspaper is unique, the common patterns between this newspaper and the others will be removed from the signature obtained in the above step. For this purpose, the notion of reference patterns is used. The reference patterns for a newspaper are formed by concatenating the POS skip-gram patterns of the other three newspapers. For example, the reference patterns of the Independent are formed by concatenating the top-250 POS skip-gram patterns of the other three newspapers. The final revised signature of a newspaper is then obtained by removing from its initial signature all patterns that also occur in its reference patterns.

## 4. RESULTS & DISCUSSION

The numbers of patterns extracted during the implementation steps are presented below.

Table 7.  Numbers of patterns extracted in the steps of implementation.

|  | Number of patterns extracted from the training data | Number of reference patterns | Number of patterns in the initial signature | Number of patterns in the revised signature |
|---|---|---|---|---|
| BBC | 55059 | 164422 | 14744 | 269 |
| The Guardian | 56744 | 162737 | 16666 | 203 |
| The Independent | 54611 | 164870 | 14290 | 203 |
| The Daily Mirror | 53067 | 166414 | 13221 | 62 |

Table 7 shows the numbers of patterns selected in each step of implementation. The numbers of patterns extracted from the training data of the four newspapers do not differ much. The number of reference patterns is, by the definition of the reference patterns, influenced by the total numbers of patterns discovered from the training data of the other three newspapers. Therefore, the number of reference patterns of the Daily Mirror is the largest, followed by the Independent, BBC, and the Guardian, that is to say, it follows a reverse trend compared with the number of patterns extracted from the training data.

As far as the number of patterns in the initial signature is concerned, the previous trend still holds: if the news articles from a newspaper are generally longer, the number of patterns that they have in common will be greater, and therefore, a greater number of patterns can be found in their initial signature.

When the reference patterns are removed from the initial signatures, the numbers of patterns that remain show a different trend. The number of patterns contained in the revised signature of BBC is the largest, followed by the Guardian, the Independent, and the Daily Mirror, which suggests that BBC has more unique patterns and the Daily Mirror has the smallest number of unique patterns. Through examination of the texts, it is found that articles from BBC contain more subheadings and links to related articles, while articles from the Daily Mirror contain fewer these types of texts, which may be one of the reasons for the greater number of unique patterns in the revised signature of BBC.

Moreover, it is noteworthy that even though the Guardian has the largest number of patterns in its initial signature and the smallest number of reference patterns to be removed from its initial signature, the number of unique patterns in its revised signature is the same as the Independent, which means that a considerable amount of patterns shared by the files in the training data of the Guardian do not have the discriminative power for telling this newspaper and the others apart.
As we are interested in finding out if different newspapers express temporal information in different ways, the patterns that involve temporal information will be examined for comparing the four newspapers in this aspect.

Since <s>, <ng>, <vg> and other tags are not directly related to temporal information and temporal relations are not completely recognized by the tagger, statistics are generated only for patterns containing tags of <EVENT> and <TIMEX3> in the following table.

Table 8. Ratios of patterns containing temporal information tags with respect to the total numbers of patterns in the revised signatures.

|  | Number of patterns containing temporal information tags | Number of patterns in the revised signature | Ratio(%) |
|---|---|---|---|
| BBC | 33 | 269 | 12.3 |
| The Guardian | 25 | 203 | 12.3 |
| The Independent | 29 | 203 | 14.3 |
| The Daily Mirror | 22 | 62 | 35.5 |

From Table 8, it can be seen that the revised signature of the Daily Mirror has a greater ratio of patterns containing temporal information tags than the other three newspapers, which seems to be contrary to our preconception. However, as temporal expressions and events generally form the essential elements in the development of a topic, if they are given explicitly, the readers can grasp the core information more quickly, which makes it easier to condense the report into a shorter article. This explanation is similar to the explanation [29] for the more frequent use of numbers in tabloids. The proportions of patterns containing temporal information tags of the other three newspapers do not differ much.

To see if the four newspapers can be distinguished by some patterns in particular, the patterns containing temporal information tags are presented below. The patterns are understood in the following way: as an example, in a pattern "2 -1 3 -1 #SUP: 5", the pattern is formed by 2 followed by 3, -1 is used to separate the items from each other, "#SUP:" denotes *support*, and the number after "#SUP:" is the value of the support of the pattern. This follows the format of the output of the SPMF interface.

Table 9.  Patterns containing temporal information tags in the revised signature of BBC.

| **BBC** | | | | | | |
|---|---|---|---|---|---|---|
| 1 | EVENT | -1 | #SUP: | 78 | | |
| 2 | vbn | -1 | EVENT | -1 | #SUP: | 18 |
| 3 | EVENT | -1 | vbd | -1 | #SUP: | 37 |
| 4 | TIMEX3 | -1 | #SUP: | 30 | | |
| 5 | EVENT | -1 | #SUP: | 74 | | |
| 6 | TIMEX3 | -1 | ng | -1 | #SUP: | 30 |
| 7 | EVENT | -1 | vbd | -1 | #SUP: | 36 |
| 8 | EVENT | -1 | #SUP: | 65 | | |
| 9 | vbn | -1 | EVENT | -1 | #SUP: | 27 |
| 10 | TIMEX3 | -1 | ng | -1 | #SUP: | 29 |
| 11 | TIMEX3 | -1 | #SUP: | 29 | | |
| 12 | EVENT | -1 | vbg | -1 | #SUP: | 36 |
| 13 | TIMEX3 | -1 | ng | -1 | #SUP: | 30 |
| 14 | EVENT | -1 | vbn | -1 | #SUP: | 35 |
| 15 | EVENT | -1 | vbd | -1 | #SUP: | 37 |
| 16 | EVENT | -1 | #SUP: | 73 | | |
| 17 | vg | -1 | EVENT | -1 | #SUP: | 67 |
| 18 | TIMEX3 | -1 | #SUP: | 30 | | |
| 19 | EVENT | -1 | vbn | -1 | #SUP: | 40 |
| 20 | vg | -1 | EVENT | -1 | #SUP: | 68 |
| 21 | EVENT | -1 | #SUP: | 70 | | |
| 22 | TIMEX3 | -1 | #SUP: | 29 | | |
| 23 | TIMEX3 | -1 | ng | -1 | #SUP: | 29 |
| 24 | EVENT | -1 | #SUP: | 47 | | |
| 25 | EVENT | -1 | #SUP: | 66 | | |
| 26 | vg | -1 | EVENT | -1 | #SUP: | 59 |
| 27 | EVENT | -1 | vbn | -1 | #SUP: | 28 |
| 28 | EVENT | -1 | vbn | -1 | #SUP: | 45 |
| 29 | EVENT | -1 | #SUP: | 64 | | |
| 30 | vbn | -1 | EVENT | -1 | #SUP: | 24 |
| 31 | vg | -1 | EVENT | -1 | #SUP: | 58 |
| 32 | vg | -1 | EVENT | -1 | #SUP: | 49 |
| 33 | vg | -1 | EVENT | -1 | #SUP: | 74 |

As can be seen from Table 9, eight of the 33 patterns in the revised signature of BBC are formed with the <TIMEX3> tag. Recall that the <TIMEX3> tag annotates explicit temporal expressions such as "today" and "on December 20, 1980". Compared with the Guardian and the Independent below, this is a much greater ratio.

Table 10. Patterns containing temporal information tags in the revised signature of the Guardian.

| The Guardian | | | | | | |
|---|---|---|---|---|---|---|
| 1 | vg | -1 | EVENT | -1 | #SUP: | 60 |
| 2 | EVENT | -1 | vbz | -1 | #SUP: | 25 |
| 3 | EVENT | -1 | #SUP: | 62 | | |
| 4 | vg | -1 | EVENT | -1 | #SUP: | 56 |
| 5 | EVENT | -1 | #SUP: | 61 | | |
| 6 | EVENT | -1 | vbd | -1 | #SUP: | 33 |
| 7 | EVENT | -1 | vbd | -1 | #SUP: | 25 |
| 8 | EVENT | -1 | vbd | -1 | #SUP: | 25 |
| 9 | vbn | -1 | EVENT | -1 | #SUP: | 21 |
| 10 | EVENT | -1 | vbn | -1 | #SUP: | 38 |
| 11 | EVENT | -1 | vbd | -1 | #SUP: | 25 |
| 12 | EVENT | -1 | vbd | -1 | #SUP: | 25 |
| 13 | EVENT | -1 | vbd | -1 | #SUP: | 31 |
| 14 | EVENT | -1 | vbd | -1 | #SUP: | 25 |
| 15 | vbn | -1 | EVENT | -1 | #SUP: | 28 |
| 16 | EVENT | -1 | vbn | -1 | #SUP: | 41 |
| 17 | EVENT | -1 | vbg | -1 | #SUP: | 37 |
| 18 | EVENT | -1 | #SUP: | 69 | | |
| 19 | EVENT | -1 | #SUP: | 68 | | |
| 20 | EVENT | -1 | vbz | -1 | #SUP: | 38 |
| 21 | vg | -1 | EVENT | -1 | #SUP: | 65 |
| 22 | EVENT | -1 | vbz | -1 | #SUP: | 38 |
| 23 | vg | -1 | EVENT | -1 | #SUP: | 80 |
| 24 | EVENT | -1 | vbg | -1 | #SUP: | 32 |
| 25 | EVENT | -1 | #SUP: | 85 | | |

In the revised signature of the Guardian, no patterns containing the tag <TIMEX3> are found, which suggests that the temporal expressions are not used in a distinctive way in the news articles of the Guardian. Among the 25 patterns, there are seven pattern comprised of the tag <EVENT> followed by the <vbd> tag.

Table 11. Patterns containing temporal information tags in the revised signature of the Independent.

| The Independent | | | | | | |
|---|---|---|---|---|---|---|
| 1 | EVENT | -1 | vbd | -1 | #SUP: | 34 |
| 2 | EVENT | -1 | vbz | -1 | #SUP: | 31 |
| 3 | EVENT | -1 | #SUP: | 60 | | |
| 4 | TIMEX3 | -1 | ng | -1 | #SUP: | 23 |
| 5 | EVENT | -1 | vbg | -1 | #SUP: | 44 |
| 6 | vg | -1 | EVENT | -1 | #SUP: | 148 |
| 7 | EVENT | -1 | vbd | -1 | #SUP: | 114 |
| 8 | EVENT | -1 | vbn | -1 | #SUP: | 61 |
| 9 | EVENT | -1 | #SUP: | 168 | | |
| 10 | TIMEX3 | -1 | #SUP: | 23 | | |
| 11 | vbn | -1 | EVENT | -1 | #SUP: | 33 |
| 12 | EVENT | -1 | vbg | -1 | #SUP: | 30 |
| 13 | EVENT | -1 | #SUP: | 63 | | |
| 14 | vg | -1 | EVENT | -1 | #SUP: | 57 |
| 15 | EVENT | -1 | vbd | -1 | #SUP: | 32 |
| 16 | vbn | -1 | EVENT | -1 | #SUP: | 19 |
| 17 | EVENT | -1 | vbz | -1 | #SUP: | 31 |
| 18 | EVENT | -1 | vbn | -1 | #SUP: | 43 |
| 19 | EVENT | -1 | #SUP: | 77 | | |
| 20 | EVENT | -1 | vbn | -1 | #SUP: | 25 |
| 21 | EVENT | -1 | #SUP: | 55 | | |
| 22 | vg | -1 | EVENT | -1 | #SUP: | 50 |
| 23 | EVENT | -1 | vbn | -1 | #SUP: | 37 |
| 24 | vbn | -1 | EVENT | -1 | #SUP: | 25 |
| 25 | EVENT | -1 | vbd | -1 | #SUP: | 35 |
| 26 | EVENT | -1 | #SUP: | 79 | | |
| 27 | vg | -1 | EVENT | -1 | #SUP: | 72 |
| 28 | EVENT | -1 | vbd | -1 | #SUP: | 32 |
| 29 | vg | -1 | EVENT | -1 | #SUP: | 51 |

The revised signature of the Independent contains two patterns formed by <TIMEX3>, which indicates that slightly more temporal expressions are used in articles from the Independent than the Guardian. However, when compared with BBC and the Daily Mirror, the temporal expressions are used much less frequently in the Independent.

Table 12.  Patterns containing temporal information tags in the revised signature of the Daily Mirror.

| The Daily Mirror | | | | | | |
|---|---|---|---|---|---|---|
| 1 | TIMEX3 | -1 | ng | -1 | #SUP: | 28 |
| 2 | EVENT | -1 | #SUP: | 50 | | |
| 3 | EVENT | -1 | vbn | -1 | #SUP: | 48 |
| 4 | vbn | -1 | EVENT | -1 | #SUP: | 34 |
| 5 | TIMEX3 | -1 | #SUP: | 28 | | |
| 6 | EVENT | -1 | vbn | -1 | #SUP: | 32 |
| 7 | TIMEX3 | -1 | ng | -1 | #SUP: | 31 |
| 8 | TIMEX3 | -1 | #SUP: | 31 | | |
| 9 | vg | -1 | EVENT | -1 | #SUP: | 44 |
| 10 | EVENT | -1 | #SUP: | 5 | | |
| 11 | vg | -1 | EVENT | -1 | #SUP: | 4 |
| 12 | TIMEX3 | -1 | #SUP: | 35 | | |
| 13 | EVENT | -1 | vbn | -1 | #SUP: | 33 |
| 14 | EVENT | -1 | #SUP: | 50 | | |
| 15 | TIMEX3 | -1 | ng | -1 | #SUP: | 35 |
| 16 | vbn | -1 | EVENT | -1 | #SUP: | 23 |
| 17 | EVENT | -1 | vbg | -1 | #SUP: | 29 |
| 18 | TIMEX3 | -1 | ng | -1 | #SUP: | 21 |
| 19 | TIMEX3 | -1 | #SUP: | 21 | | |
| 20 | vg | -1 | EVENT | -1 | #SUP: | 46 |
| 21 | EVENT | -1 | vbz | -1 | #SUP: | 26 |
| 22 | EVENT | -1 | #SUP: | 48 | | |

Among the 22 patterns in the revised signature of the Daily Mirror, eight are patterns containing the tag of <TIMEX3>, which is an indicator that the Daily Mirror tends to use more explicit temporal expressions in its news articles than the other newspapers. In combination with Table 8, it may be concluded that the news articles of the Daily Mirror, generally described as a tabloid, contain a greater proportion of temporal information and the temporal information is expressed with explicit temporal expressions, which makes it easier to convey the essentials of a piece of news within limited space. In contrast, articles from the Guardian and the Independent are more likely to convey temporal information implicitly. BBC lies in the middle of this spectrum of explicitness and implicitness.

In the authorship attribution step, an experiment following the method used in [1] is performed. The accuracies for BBC, the Guardian, the Independent and the Daily Mirror are 90.7%, 6.7%, 0% and 0%, respectively. In the experiment [1], the success ratio for the classification task on Catharine Trail is equal to zero under the same setting as in the present experiment. It is explained in [1] that some authors are harder to identify because some of them attempt to hide their identities with deliberate variations in style, and therefore the patterns in their signatures are heterogeneous, or because some authors write about the daily routine life, which makes their writing share much in common with the other authors. Except for BBC, all the other newspapers are not classified with high accuracy, which may be attributed to the fact that BBC contains more subheadings and links to related articles and this feature can be more easily characterized by skip-gram patterns.

## 5. CONCLUSIONS & FUTURE WORK

In this study, the focus is to find if different newspapers express temporal information in different ways and how they differ in terms of the amount of temporal information and the specific patterns containing temporal information tags.

From the perspective of the number of patterns in the revised signatures, it can be seen that the number of patterns contained in the revised signature of BBC is the largest, followed by the Guardian, the Independent and the Daily Mirror, which suggests that BBC has more unique patterns and the Daily Mirror has the smallest number of unique patterns. This may be attributed in part to the fact that articles from BBC contain more subheadings and inserted titles of articles related to a topic than the other newspapers while articles from the Daily Mirror, constrained by space, normally do not contain these types of texts. The POS skip-gram patterns can capture this aspect.

As articles from the Guardian, generally described as a broadsheet, are longer, more patterns are mined under the same setting of the algorithm and therefore, the initial signature of the Guardian contains the greatest number of patterns. However, even though the reference patterns of the Guardian are the smallest in number, when the initial signature of the Guardian is revised by moving the reference patterns, the patterns that remain in its revised signature are not the largest in number, which means that a considerable amount of patterns shared by the files in the training data of the Guardian do not have the discriminative power for telling this newspaper and the others apart and the language use of the Guardian is possibly not so distinctive from the other newspapers in the dataset.

From the perspective of the ratio of patterns containing temporal information tags to the number of patterns in the revised signatures, it can be seen that the revised signature of the Daily Mirror which is generally described as a tabloid has a greater ratio of patterns containing temporal information tags than the other three newspapers, which may run counter to some preconceptions. However, since temporal expressions and events generally form the essential elements of a piece of news, if they are given explicitly, the readers can grasp the core information more quickly, which makes it easier to condense a report into a shorter article. This result is consistent with the research [29] which shows that tabloids use more numbers than broadsheets for similar reasons.
As far as the specific patterns containing temporal information tags are concerned, it can be concluded that articles from the Daily Mirror, typically described as a tabloid, contain a greater proportion of temporal information and the temporal information is expressed more frequently with explicit temporal expressions than the other newspapers.

Due to the constraint of the current version of TTK, only tags of <EVENT> and <TIMEX3> are analyzed and studied. If the temporal information can be annotated fully, it is likely that a more complete picture of the different ways in expressing temporal information can be obtained.
In future work, experiment can be carried out using other methods for the authorship attribution task. Meanwhile, different datasets can be used for testing the results of this study, and how adjusting the parameter setting of the pattern mining algorithm influences the result remains a question that needs further investigation.

# REFERENCES

[1]   Pokou, Y. J. M., Fournier-Viger, P., & Moghrabi, C. (2016) "Authorship attribution using small sets of frequent part-of-speech skip-grams", *The Twenty-Ninth International Flairs Conference*.

[2]   Crystal, D. & Davy, D. (2016) *Investigating English Style*, Routledge.

[3]   Fowler, R. (2013) *Language in the News: Discourse and Ideology in the Press,* Routledge.

[4]   Bagnall, N. (1993) *Newspaper language,* Routledge.

[5]   Timuçin, M. (2010) "Different language styles in newspapers: An investigative framework", *Journal of Language and Linguistic Studies*, Vol. 6, No. 2, pp104–126.

[6]   Guthrie, D., Allison, B., Liu, W., Guthrie, L., & Wilks, Y. (2006) "A closer look at skip-gram modelling", *LREC*, Vol. 6, pp 1222–1225.

[7]    Pustejovsky, J., Lee, K., Bunt, H., & Romary, L. (2010) "ISO-TimeML: An International Standard for Semantic Annotation", *LREC*, Vol. 10, pp394–397.

[8]   Allen, J. F. (1983) "Maintaining knowledge about temporal intervals", *Communications of the ACM*, Vol. 26, No.11, pp832–843.

[9]   Pustejovsky, J., Ingria, R., Saurí, R., Castaño, J. M., Littman, J., Gaizauskas, R. J., Setzer, A., Katz, G., & Mani, I. (2005) "The Specification Language TimeML" [Online]. Available: http://www.timeml.org/timeMLdocs/timeMLspec.pdf.

[10]  Wang, W., Kreimeyer, K., Woo, E. J., Ball, R., Foster, M., Pandey, A., Scott, J., & Botsis, T. (2016) "A new algorithmic approach for the extraction of temporal associations from clinical narratives with an application to medical product safety surveillance reports", *Journal of Biomedical Informatics*, Vol. 62, pp78– 89.

[11]  Strötgen, J. & Gertz, M. (2010) "Heideltime: High quality rule-based extraction and normalization of temporal expressions", *Proceedings of the 5th International Workshop on Semantic Evaluation*, pp321–324.

[12]  Chang, A. X. & Manning, C. D. (2012) "Sutime: A library for recognizing and normalizing time expressions", *LREC*, Vol. 2012, pp3735–3740.

[13]  Lapata, M. & Lascarides, A. (2004) "Inferring sentence-internal temporal relations", *Proceedings of the Human Language Technology Conference of the North American Chapter of the Association for Computational Linguistics: HLT-NAACL*, pp153–160.

[14]  Yang, Y.-L., Lai, P.-T., & Tsai, R. T.-H. (2014) "A hybrid system for temporal relation extraction from discharge summaries", *International Conference on Technologies and Applications of Artificial Intelligence*, pp379–386.

[15]  Chang, Y.-C., Dai, H.-J., Wu, J. C.-Y., Chen, J.-M., Tsai, R. T.-H., & Hsu, W.-L. (2013) "Tempting system: a hybrid method of rule and machine learning for temporal relation extraction in patient discharge summaries", *Journal of Biomedical Informatics*, Vol. 46, ppS54–S62.

[16]  Fournier-Viger, P., Lin, J. C.-W., Kiran, R. U., Koh, Y. S., & Thomas, R. (2017) "A survey of sequential pattern mining", *Data Science and Pattern Recognition*, Vol. 1, No. 1, pp54–77.

[17]  Agrawal, R., Imieliński, T., & Swami, A. (1993) "Mining association rules between sets of items in large databases", *Proceedings of the 1993 ACM SIGMOD International Conference on Management of Data*, pp207–216.

[18]  Maylawati, D., Aulawi, H., & Ramdhani, M. (2018) "The concept of sequential pattern mining for text", *IOP Conference Series: Materials Science and Engineering*, Vol. 434, No.012042, doi:10.1088/1757-899X/434/1/012042.

[19]  Hoonlor, A. (2011) *Sequential patterns and temporal patterns for text mining*. Rensselaer Polytechnic Institute.

[20]  Hirate, Y. & Yamana, H. (2006) "Sequential pattern mining with time intervals", *Pacific-Asia Conference on Knowledge Discovery and Data Mining*, pp775–779.

[21]  Laxman, S. &  Sastry, P. S. (2006) "A survey of temporal data mining", *Sadhana*, Vol. 31, No. 2, pp173–198.

[22]  Yen, S.-J. & Lee, Y.-S. (2013) "Mining non-redundant time-gap sequential patterns", *Applied Intelligence*, Vol. 39, No.4, pp727–738.

[23]  Iqbal, F., Binsalleeh, H., Fung, B. C., & Debbabi, M. (2010) "Mining writeprints from anonymous e-mails for forensic investigation", *Digital Investigation*, Vol. 7, No. 1-2, pp56–64.

[24]  Verhagen, M. & Pustejovsky, J. (2012) "The Tarsqi Toolkit", *LREC*, pp2043–2048.

[25]  Schmid, H. (1995) "Improvements in part-of-speech tagging with an application to German", *Proceedings of the ACL SIGDAT-Workshop*, pp47–50.

[26]  Fournier-Viger, P., Gomariz, A., Gueniche, T., Mwamikazi, E., & Thomas, R. (2013) "TKS: efficient mining of top-k sequential patterns", *International Conference on Advanced Data Mining and Applications*, pp109–120.

[27]  Ayres, J., Flannick, J., Gehrke, J., & Yiu, T. (2002) "Sequential pattern mining using a bitmap representation", *Proceedings of the Eighth ACM SIGKDD International Conference on Knowledge Discovery and Data Mining*, pp429–435.

[28]  Fournier-Viger, P., Gomariz, A., Gueniche, T., Soltani, A., Wu, C.-W., & Tseng, V. S. (2014) "Spmf: a java open-source pattern mining library", *The Journal of Machine Learning Research*, Vol. 15, No. 1, pp3389–3393.

[29]  Li, Y., Zhang, D., & Wanyi, D. (2014) "A case analysis of lexical features in english broadsheets and tabloids", *International Journal of English Linguistics*, Vol. 4, No. 4, p115-122.

**AUTHORS**

**Yingxue Fu** is a Ph.D. candidate at the University of St Andrews, UK. She studied under the program of M.Phil. in Speech and Language Processing at Trinity College Dublin in 2019-2020.

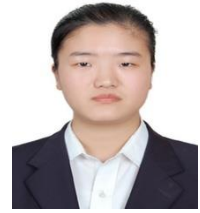

**Dr Elaine Uí Dhonnchadha** is Assistant Professor in Computational Linguistics and M.Phil Coordinator for the Centre for Language and Communication Studies. Prior to joining Trinity College, Elaine worked as a researcher in Institiúid Teangeolaíochta Éireann (The Linguistics Institute of Ireland), as a Lecturer in Dublin City University and as a Systems Analyst and Programmer in a number of software development companies.

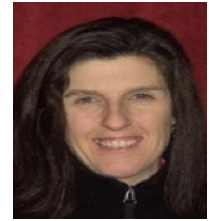